\documentclass[british,a4paper,conference]{llncs}
\usepackage{lmodern}
\usepackage[latin9]{inputenc}
\usepackage{array}
\usepackage{booktabs}
\usepackage{units}
\usepackage{amsmath}
\usepackage{amssymb}
\usepackage{graphicx}

\makeatletter

\providecommand{\tabularnewline}{\\}

\usepackage{balance}
\usepackage{flushend}

\makeatother

\usepackage{babel}
\begin{document}

\title{Stabilizing Linear Prediction Models using Autoencoder}

\author{Shivapratap Gopakumar, Truyen Tran, Dinh Phung, Svetha Venkatesh}

\institute{Center for Pattern Recognition and Data Analytics, Deakin University,
Australia\\
\{sgopakum, truyen.tran, dinh.phung, svetha.venkatesh\}@deakin.edu.au}

\maketitle
\global\long\def\xb{\boldsymbol{x}}
\global\long\def\wb{\boldsymbol{w}}
\global\long\def\ub{\boldsymbol{u}}
\global\long\def\hb{\boldsymbol{h}}
\global\long\def\bb{\boldsymbol{b}}
\global\long\def\zb{\boldsymbol{z}}
\global\long\def\dataset{\mathcal{D}}
\global\long\def\realset{\mathbb{R}}

\begin{abstract}
To date, the instability of prognostic predictors in a sparse high
dimensional model, which hinders their clinical adoption, has received
little attention. Stable prediction is often overlooked in favour
of performance. Yet, stability prevails as key when adopting models
in critical areas as healthcare. Our study proposes a stabilization
scheme by detecting higher order feature correlations. Using a linear
model as basis for prediction, we achieve feature stability by regularizing
latent correlation in features. Latent higher order correlation among
features is modelled using an autoencoder network. Stability is enhanced
by combining a recent technique that uses a feature graph, and augmenting
external unlabelled data for training the autoencoder network. Our
experiments are conducted on a heart failure cohort from an Australian
hospital. Stability was measured using Consistency index for feature
subsets and signal-to-noise ratio for model parameters. Our methods
demonstrated significant improvement in feature stability and model
estimation stability when compared to baselines.

\end{abstract}

\section{Introduction}
Healthcare data is expected to increase by fifty-fold in the coming
years \cite{raghupathi2014big}. While the direction of current machine
learning research is to handle such data, clinical interpretability
of models is often overlooked. In healthcare, interpretability is
the ability of the model to explain the reason behind prognosis. Such
models identify a small subset of strong features (predictors) from
available data, and rank them according to their predictive power
\cite{zhou2013patient}. This act of feature selection and ranking
need to be stable in the face of data re-sampling to ensure clinical
adoption. Nonetheless, the nature of clinical data introduces several
challenges.

For a particular condition, training data derived from electronic
medical records (EMR) usually consist of small number of cases with
a large number of events. Most of these events have high correlation
with each other. As example, emergency admission events will be correlated
with ward transfers, diagnosis of co-occurring diseases (heart failure
and diabetes) will have high correlation, pathological measurements
(amount of Sodium and Potassium in the body) will be related. To avoid
over-fitting, such data require sparse methods in feature selection
and learning \cite{Ye2012a}. But sparsity in correlated features
causes instability and results in non-reproducible models \cite{Austin2004a,lin2013high}.

In the presence of correlated features, automatic feature selection
using lasso has proven to be unstable for linear \cite{zhao2006model}
and survival models \cite{lin2013high}. Recent studies propose adapting
lasso to acknowledge feature correlations. Such correlations or groupings
can be identified using cluster analysis \cite{au2005attribute,ma2007supervised,park2007averaged}
or density estimation \cite{Yu2008}. Group lasso and its variants
have been introduced for scenarios where feature groupings are predefined
\cite{Yuan2006,jacob2009group}. When the features are ordered, or
at least have a specification of the nearest neighbour of each feature,
Tibshirani~et~al. proposed  fused lasso to perform feature grouping
\cite{Tibshirani2005}. In contrast, elastic net regularization forces
sharing of statistical weights in correlated features without imposing
any preconditions on data \cite{zou2005regularization}. When applied
to clinical prediction, elastic net regularization proved superior
to lasso for prostate cancer dataset \cite{simon2011regularization}.
Another approach to stabilize a sparse model is by additional regularization
using graphs, where the nodes are features and edges represent relationships
\cite{Sandler2009}. This strategy has been successfully applied in
bioinformatics, where feature interactions have been extensively documented
and stored in online databases \cite{Li2008,Cun2013}. In clinical
setting, recent studies have used covariance graph \cite{kamkar2015exploiting}
and hand crafted feature graph using semantic relations in ICD-10\footnote{http://apps.who.int/classifications/icd10}
diagnosis codes and intervention codes\footnote{https://www.aihw.gov.au/procedures-data-cubes/}
\cite{gopakumar_et_al_jbhi} as solutions to the instability problem.
However, these studies did not consider higher order correlations,
and lacked capability to automatically learn feature groupings. 
\begin{figure}[t]
\centering{}\includegraphics[width=0.5\columnwidth]{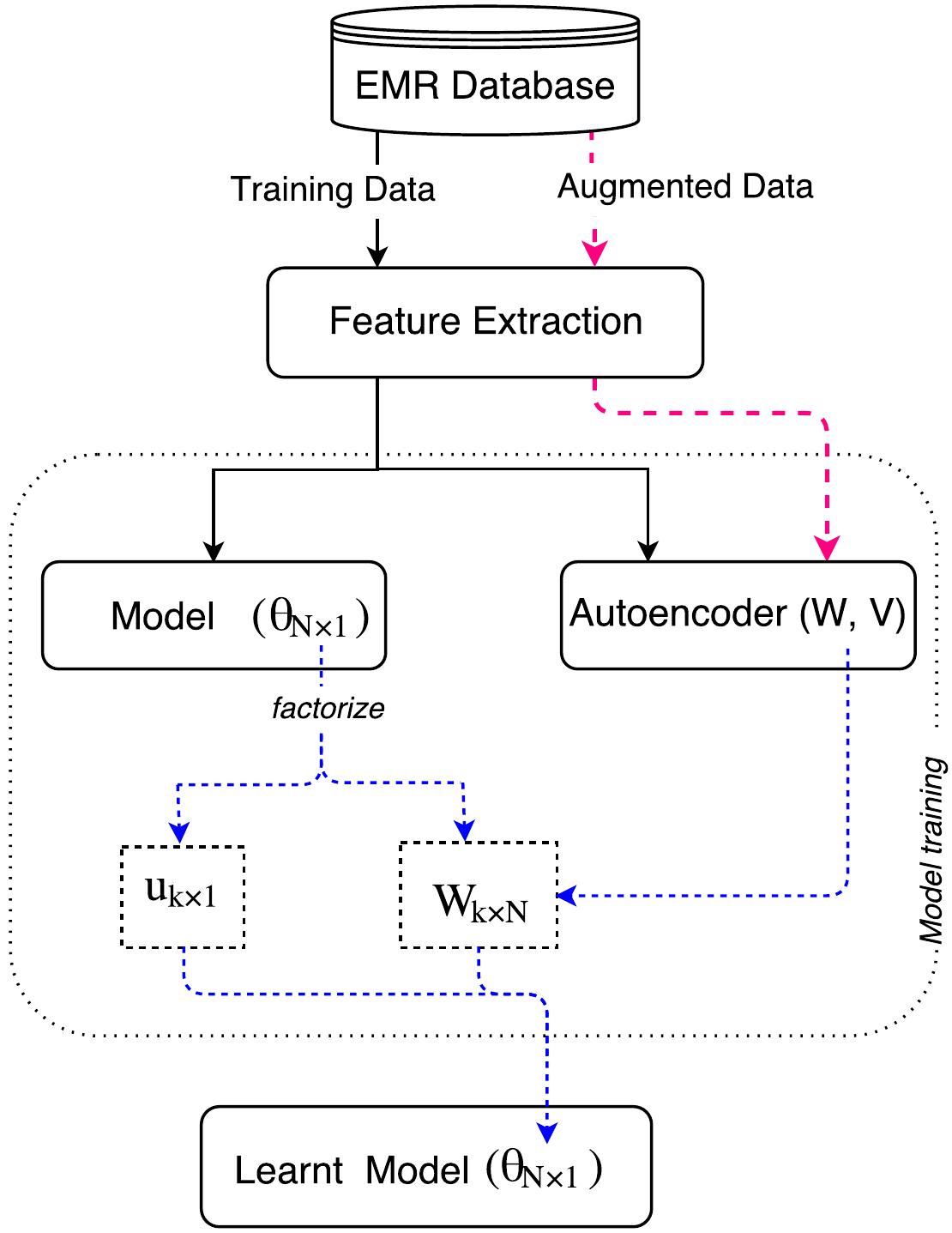}\caption{The work-flow diagram of our framework for deriving autoencoder stabilized
prediction model from EMR. The model parameter $\theta$ is factorized
into a lower dimensional vector $u$ and high dimensional matrix $W$.
The $W$ matrix is jointly modelled as encoding weights in an autoencoder
network and is used to regularize the prediction model. \label{fig:work-flow-diagram}}
\end{figure}

In this paper, we propose a novel methodology to stabilize a sparse
high dimensional linear model using recent advances in deep learning
and self-taught learning \cite{raina2007self}. We propose that the
linear model parameter $\theta$ is a combination of a lower dimensional
vector $u$, and a high dimensional matrix $W$, where $W$ encapsulates
the feature correlations. By modelling $W$ as the encoding weights
of an autoencoder network, we capture higher order feature correlations
in data. The workflow diagram of our method is illustrated in Fig.\ref{fig:work-flow-diagram}.

To minimize variance in feature subsets and parameter estimation,
we introduce three regularizers for our sparse linear model: 1) autoencoder
derived from training cohort, 2) combination of autoencoder and feature
graph derived from training cohort, 3) combination of feature graph
derived from training cohort and autoencoder derived from augmenting
an external cohort to training data. This process of augmenting external
data to autoencoder training results in more robust estimation of
higher order correlation matrix $W$.

We conducted our experiments on $1,885$ heart failure admissions
from an Australian hospital. The augmented external data consisted
of $2,840$ diabetic admissions. Feature stability was measured using
consistency index \cite{Kuncheva2007}. Parameter estimation stability
was measured using signal-to-noise ration (SNR). Our proposed stabilization
methods demonstrated significantly higher stability when compared
with the baselines.Our contribution is in understanding the need
for stable prediction, when much research has been dedicated to improving
performance. For critical applications like healthcare, where data
is sparse and redundant, stable features and estimates are necessary
to lend credence to the model and its performance.

\section{Framework}
\global\long\def\xb{\boldsymbol{x}}
\global\long\def\wb{\boldsymbol{w}}
\global\long\def\ub{\boldsymbol{u}}
\global\long\def\hb{\boldsymbol{h}}
\global\long\def\bb{\boldsymbol{b}}
\global\long\def\zb{\boldsymbol{z}}

Sparse generalized linear models take the form $f(x)=\theta^{T}x$
subject to $\sum_{i=1}^{N}|\theta_{i}|\le\alpha$, where $\theta\in\mathbb{R}^{N}$
is the model parameter derived from data: $\xb\in\mathbb{R}^{N}$.
Here, $\alpha$ is the sparsity controlling parameter, typically enforced
using lasso regularization \cite{Tibshirani1996}. More formally,
let $\mathcal{D}=\left\{ \xb_{m},\,y_{m}\right\} _{m=1}^{M}$ denote
the training data, where $\xb_{m}\in\mathcal{R}^{N}$ denotes the
high dimensional feature vector of data instance $m$, and $y_{m}$
is the outcome (for example, the occurrence of future readmission).
If $\mathcal{L}(\theta|\mathcal{D})$ is a linear loss, we have:

\begin{equation}
\mathcal{L}_{\text{lasso}}=\frac{1}{M}\mathcal{L}(\theta|\mathcal{D})+\alpha\sum_{i=1}^{N}|\theta_{i}|\label{eq:sparse-logistic-loss}
\end{equation}
where $\alpha>0$ controls the sparsity of the model parameters. The
lasso regularization forces the weights of weak parameters towards
zero. However, enforcing sparsity on high-dimensional, highly correlated
and redundant data, as derived from EMR causes the following problems.

First, lasso regularization randomly chooses one feature from a correlated
pair. The EMR data has a high degree of correlation and redundancy
in hospital recorded events. Since lasso favours stronger predictors,
each feature from a highly correlated pair will only have about 50\%
chance in being selected during every training run. Second, in clinical
scenario, a large collection of features could be weakly predictive
for a condition or medical event. Lasso could ignore such feature
groups due to low selection probabilities \cite{Meinshausen2010}. 

This sparsity-stability predicament can be resolved by forcing correlated
features to have similar weights. The traditional approach is to modify
lasso regularization with elastic net \cite{zou2005regularization}
as:

\[
\mathcal{L}_{\text{elastic net}}=\frac{1}{M}\mathcal{L}(\theta|\mathcal{D})+\alpha\left(\lambda_{\text{en}}\sum_{i}|\theta_{i}|+(1-\lambda_{\text{en}})\sum_{i}\theta_{i}^{2}\right)
\]
Here, the ridge regression term $\sum_{i}\theta_{i}^{2}$ encourages
correlated feature pairs to have similar weights, and $\lambda_{\text{en}}\in[0,1]$
balances the contribution of lasso term $\sum_{i}|\theta_{i}|$ and
ridge regression term $\sum_{i}\theta_{i}^{2}$. However, elastic
net forces the weights to be equally small, resulting in lesser sparsity. 

Another recent work introduces a feature graph built from hierarchical
relations in ICD-10 diagnosis codes and intervention codes \cite{gopakumar_et_al_jbhi}.
A Laplacian of this feature graph: $\mathbf{L}$, is used to ensure
weight sharing between related features as:

\begin{equation}
\mathcal{L}_{\text{feature graph}}=\mathcal{L}_{\text{lasso}}+\frac{1}{2}\,\lambda_{\text{fg}}(\theta^{T}\mathbf{L}\theta)\label{eq:feature-graph}
\end{equation}
We propose to automatically learn higher order correlations in data.

\subsection{Correlation by Factorization in Linear Models}

To model higher order correlations in data, we begin by decomposing
model parameter $\theta$ into a lower order vector and a high dimensional
matrix as: $\theta_{\text{N}\times1}=W_{k\times\text{N}}^{T}\,u_{k\times1}$,
where $k\ll\text{N}$. This factorization offers several advantages.
The lower dimensionality of $u$ makes it more easier to learn and
more stable to data variations. The $W$ captures higher order correlations
that be modelled using different auxiliary tasks. Greater number of
tasks ensure better solution, since there are more constraints. 

As a concrete example for generalized linear models, we work on binary
prediction using logistic regression. The modified logistic loss function
$\mathcal{L}(\theta|\mathcal{D})$ using $u$ and $W$ becomes:

\begin{eqnarray}
\mathcal{L}_{\text{logit}}(u,W\,|\mathcal{D}) & = & \text{log}(1+\text{exp}(-yu^{T}W\xb))\label{eq:logistic-model}\\
 & = & \text{log}(1+\text{exp}(-yu^{T}z))\nonumber 
\end{eqnarray}
where $y\in\pm1$ represents the data label\footnote{We ignore the bias parameter for simplicity}.
Notice that $z=W\xb$ is a data transformation from N dimensions to
the smaller $k$ dimension. To learn $W$, we need to choose a competent
auxiliary task. We model $W$ as the encoding weights of a classical
autoencoder derived from the same data $\mathcal{D}$. 

\subsection{Learning Higher Order Correlations using Autoencoder}

An autoencoder is a neural network that learns by minimizing the reconstruction
error using back-propagation \cite{bengio2009learning}. The learning
process is unsupervised, wherein the model learns the useful properties
of the data. An autoencoder network consists of two components: (1)
An \emph{encoder function }that maps the input data $\xb\in\mathbb{R}^{N}$
as: $h(x)=\sigma(W\xb+b_{W})$ , where $\sigma$ can be any non-linear
function (for e.g., the sigmoid function) and $W,b_{W}$ are the weights
and bias of the hidden layer (2) A \emph{decoder function} that attempts
to reconstruct the input data as: $\tilde{x}=Vh+b_{V}$, where $V,b_{V}$
are the weights and bias of the output layer. The loss function is
modelled as the reconstruction error:

\begin{equation}
\mathcal{L_{\text{AE}}}(W,V,b_{W,}b_{V}|\mathcal{D})=\frac{1}{2N}||x-b_{V}-V\sigma(Wx+b_{W})||_{2}^{2}\label{eq:autoencoder}
\end{equation}
Once trained, evaluating a feed forward mapping using the encoder
function gives a latent representation of the data. When the number
of hidden units is significantly lesser than the input layer, $W$
encapsulates the higher order correlations among features.

We propose to regularize our sparse linear model in (\ref{eq:logistic-model})
using the autoencoder framework in (\ref{eq:autoencoder}). The joint
loss function becomes:

\begin{eqnarray}
\mathcal{L}_{\text{model}}(u,W,V,b_{W},b_{V}|\mathcal{D}) & = & \mathcal{L}_{\text{logit}}(u,W\,|\mathcal{D})+\label{eq:basic-model-loss}\\
 & + & \alpha\underset{_{i}}{\varSigma}|\underset{_{k}}{\varSigma}W_{ik}^{T}u_{k}|\nonumber \\
 & + & \lambda_{\text{AE}}\mathcal{\,L_{\text{AE}}}(W,V,b_{W},b_{V}|\mathcal{D})\nonumber \\
 & + & \lambda_{\ell2}\left(W^{2}+V^{2}+b_{W}^{2}+b_{V}^{2}\right)\nonumber 
\end{eqnarray}
where $\alpha>0$ is the lasso regularization parameter which ensures
weak $\theta_{i}=\underset{_{k}}{\varSigma}W_{ik}^{T}u_{k}$ are driven
to zero. While $\lambda_{\text{AE}}$ controls the amount of regularization
due to higher order correlation, $\lambda_{\ell2}$ controls overfitting
in autoencoder. The loss function in (\ref{eq:basic-model-loss})
is non-convex. We propose two extensions to our model.

\subsubsection{Augmenting Feature Graph regularization}

While autoencoders can be used to find automatic feature grouping,
we can also exploit the predefined associations in patient medical
records. For example, diseases or conditions reoccurring over multiple
time-horizons should be assigned similar importance \cite{tran2013integrated}.
Also, the ICD-10 diagnosis and procedure codes are hierarchical in
nature \cite{Truyen2014ordinal_kais,gopakumar_et_al_jbhi}. We build
a feature graph using these associations (as in (\ref{eq:feature-graph}))
and use it to further regularize our model in (\ref{eq:basic-model-loss})
as:

\begin{eqnarray*}
\mathcal{L}_{\text{model-fg}}(u,W,V,b_{W},b_{V}|\mathcal{D}) & = & \mathcal{L}_{\text{model}}(u,W,V,b_{W},b_{V}|\mathcal{D})\\
 & + & \frac{1}{2}\,\lambda_{\text{fg}}\left[(u^{T}W)\mathbf{\,L}\,(W^{T}u)\right]
\end{eqnarray*}

\subsubsection{Augmenting External data for Autoencoder learning}

The encoding weights $W$ in (\ref{eq:autoencoder}) can be estimated
from multiple sources. For example, in this paper, we propose to augment
the current training data $\mathcal{D}$ (for example: heart failure
cohort) with another cohort containing the same features (say, diabetic
cohort). Training the autoencoder network on this augmented data will
result in more robust estimation of $W$.

\section{Experiments}

The feature stability and model stability of our proposed framework
is evaluated on heart failure (HF) cohort from Barwon Health\footnote{Ethics approval was obtained from the Hospital and Research Ethics
Committee at Barwon Health (number 12/83) and Deakin University.}, a regional hospital in Australia serving more than 350,000 residents.
The Autoencoder learning was augmented with diabetes (DB) cohort from
the same hospital. We mined the hospital EMR database for retrospective
data for a period of 5 years (Jan 2007 to Dec 2011), focusing on emergency
and unplanned admissions of all age groups. Inpatient deaths were
excluded. All patients with at least one ICD-10 diagnosis code I50
were included in the HF cohort. The DB cohort contained all patients
with at least one diagnosis code between E10-E14. This resulted in
$1,885$ heart failure admissions and $2,840$ diabetic admissions.
Table~\ref{tab:CohortDetails} shows the details of both cohorts.
\begin{table}[h]
\begin{centering}
\caption{Characteristics of heart failure and diabetes cohort.\label{tab:CohortDetails}}
\par\end{centering}
\centering{}%
\begin{tabular}{>{\raggedleft}m{0.25\columnwidth}>{\centering}m{0.15\columnwidth}>{\centering}m{0.15\columnwidth}>{\centering}m{0.15\columnwidth}}
\hline 
 & \multicolumn{2}{c}{\textbf{Heart Failure (HF)}} & \textbf{Diabetes (DB)}\tabularnewline
\hline 
 & Derivation & Validation & (Augmented data)\tabularnewline
\hline 
\noalign{\vskip\doublerulesep}
Admissions & 1,415 & 369 & 2,840\tabularnewline
\noalign{\vskip\doublerulesep}
Unique patients & 1,088 & 317 & 1,716\tabularnewline
\noalign{\vskip\doublerulesep}
Gender: &  &  & \tabularnewline
\qquad{}\qquad{}Male  & 541~(49.7\%) & 155~(48.9\%) & 908~(52.9\%)\tabularnewline
\qquad{}\qquad{}Female  & 547~(50.2\%) & 162~(51.1\%) & 808~(47.1\%)\tabularnewline
\noalign{\vskip\doublerulesep}
Mean age (years) & 78.3  & 79.4 & 57.1\tabularnewline
Mean Length of Stay & 5.2 days & 4.5 days & 4.1 days\tabularnewline
Total Features  & \multicolumn{2}{c}{3,338} & 6,711\tabularnewline
\noalign{\vskip\doublerulesep}
Common features & \multicolumn{3}{c}{558}\tabularnewline
\hline 
\end{tabular}
\end{table}

The different features in EMR database (diagnosis, medications, treatments,
procedures, lab results) were extracted using a one-sided convolutional
filter bank introduced in \cite{tran2013integrated}. The feature
extraction process resulted in $3,338$ features for HF and $6,711$
features for DB cohort. A total of $558$ features were common to
both cohorts (Table.~\ref{tab:CohortDetails}).

\subsection{Models and Baselines}

From this data, we derive a lasso regularized logistic regression
model to predict heart failure readmissions in 6 months. We force
lasso to consider higher order correlations in data by using the following
three regularization schemes:

\subsubsection{Lasso-Autoencoder}

The linear model is regularized by an autoencoder derived from HF
cohort.

\subsubsection{Lasso-Autoencoder-Graph}

We construct a feature graph from $3,338$ features in HF cohort as
in \cite{gopakumar_et_al_jbhi}, and use it to further regularize
the Lasso-Autoencoder model.

\subsubsection{AG-Lasso-Autoencoder-Graph}

AG denotes augmented data used to train the autoencoder. To estimate
W, we used DB cohort augmented to the HF cohort. Training data consisted
of $558$ features common to both HF and DB. The sparse prediction
model was built from common features in HF cohort, and regularized
using a HF-based feature graph and autoencoder from augmented data. 

We compare the stability of our proposed regularization methods with
the following baselines: 1) pure lasso 2) elastic net and 3) recently
introduced feature graph regularization (as in \cite{gopakumar_et_al_jbhi})
. 

\subsection{Temporal Validation}

The training and testing data were separated in time and patients.
Patients discharged before September 2010 were included in training
set. The validation set consisted of new admissions from September
2010 to December 2011. Model performance was measured using AUC (area
under the ROC curve) based on Mann-Whitney statistic. A pre-defined
threshold was used (chosen to maximize the F-score) to predict readmissions. 

\subsection{Measuring Stability}

Variability in data resampling was simulated using bootstraps. We
trained all models using $500$ bootstraps of randomly sampled training
data. At the end of each bootstrap, the features selected by a model
were ranked based on importance. Feature importance was calculated
as the product of mean feature weights across all bootstraps and feature
standard deviation in the training data. The top $k$ ranked features
were collected to form a list of feature subsets: $S=\{S_{1},\,S_{2},\,\cdots,\,S_{500}\}$,
where $|S_{i}|=k$. 

We used consistency index \cite{Kuncheva2007} to measure pairwise
feature subset stability. For a pair of subsets $(S_{i},\,S_{j})$,
with length $k$ selected from a total of $d$ features, consistency
index becomes:

\[
\text{CI}(S_{i},\,S_{j})=\frac{rd-k^{2}}{k(d-k)}
\]
where $r=|S_{i}\cap S_{j}|$. The overall stability score is the average
of pairwise consistency index among all pairs. This stability score
is bounded in $[-1,\,+1]$, with $-1$ for no overlap, $0$ for independently
drawn subsets, and $+1$ complete overlap of two subsets. The score
calculation is monotonous and corrects overlap due to chance \cite{Kuncheva2007}.

The stability of estimated model parameters was measured using signal-to-noise
ratio (SNR). The variance in estimation of feature $i$ can be calculated
as:

\[
\text{SNR}(i)=\nicefrac{w_{i}}{\sigma_{i}}
\]
where $w_{i}$ is the mean feature weight across bootstraps for feature
$i$, and $\sigma_{i}$ is its standard deviation.

\section{Results}
In this section, we demonstrate the effect of autoencoder regularization
on model performance and stability, and compare with our baselines.
The prediction models for heart failure readmission were derived from
$3,338$ features extracted from hospital database. The self taught
learning stage during autoencoder training used an augmented $2,840$
diabetic admissions with $558$ features that were common in both
cohorts. A grid search for the best hyper-parameter setting resulted
in $\alpha=.001$, $\lambda_{\text{en}}=.01$, $\lambda_{\text{graph}}=.03$
for the baseline models, and $\alpha=.005$, $\lambda_{\text{AE}}=3000$,
$\lambda_{\text{graph}}=0.3$ for our autoencoder regularized models. 

\subsection{Capturing Higher Order Correlations}

The efficacy of autoencoder network to model higher order correlations
was verified by comparing the correlation matrices of raw data and
data from the encoding layer.  The autoencoder derived correlation
matrix was denser (matrix mean = $0.19$) than the correlation matrix
for raw data (matrix mean = $0.05$).

\subsection{Effect on Model Sparsity }

\begin{figure}
\centering{}\includegraphics[width=0.4\paperwidth]{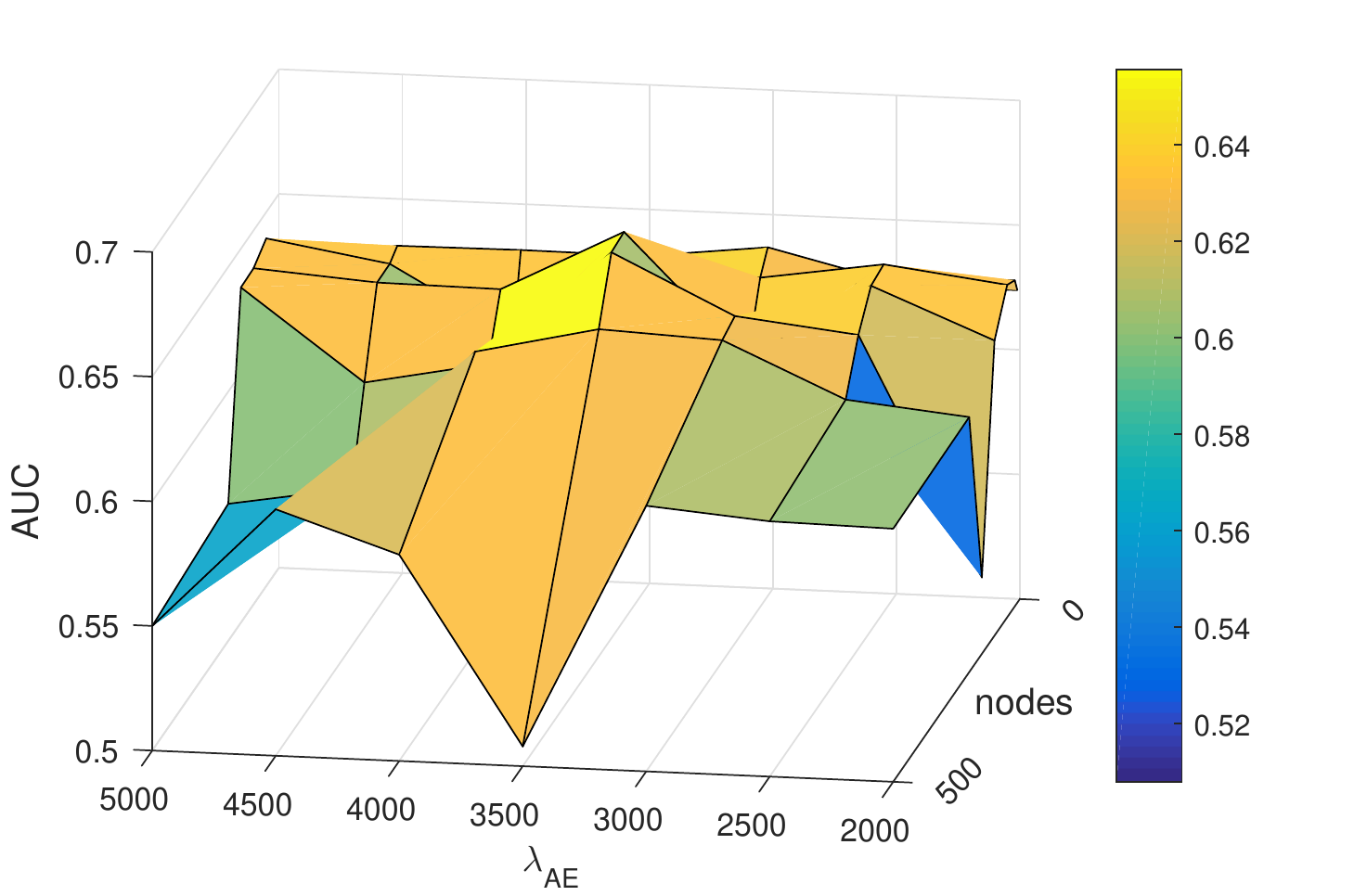}\caption{Effect of number of hidden units (nodes) and autoencoder penalty ($\lambda_{\text{AE}}$)
on AUC. Lasso parameter fixed at $\alpha=.005$ \label{fig:hyperparams-AUC}}
\end{figure}

Table ~\ref{tab:Comparison-summary} provides a summary of the effects
of stabilization schemes on model sparsity. Autoencoder regularization
resulted in sparser models with no loss in performance.
\begin{table}[h]
\centering{}\caption{Effect of stabilization methods on model sparsity \label{tab:Comparison-summary}}
\begin{tabular}{rc}
\toprule 
Regularization & ~~Features Selected (\%)\tabularnewline
\midrule
Lasso & 550~~(16.5~\%)\tabularnewline
\midrule
Elastic Net & 753~~(22.6~\%)\tabularnewline
\midrule
Lasso-Graph & 699~~(20.9~\%)\tabularnewline
\midrule
Lasso-Autoencoder & 513~(15.4~\%)\tabularnewline
\midrule
Lasso-Autoencoder-Graph & 503~(15.1~\%)\tabularnewline
\midrule
AG-Lasso-Autoencoder-Graph & 412~(12.3~\%)\tabularnewline
\bottomrule
\end{tabular}
\end{table}
 Model performance was measured using area under the ROC curve (AUC).
For autoencoder regularization, AUC critically depended on the choice
of autoencoder penalty ($\lambda_{\text{AE}}$) and number of hidden
units (see Fig.~\ref{fig:hyperparams-AUC}). 
 A maximum AUC of 0.65 was obtained for AG-Lasso-Autoencoder-Graph
model with 20 hidden units and hyperparameters as $\alpha=.005$,
$\lambda_{\text{en}}=.03$, $\lambda_{\text{graph}}=.3$, $\lambda_{\text{AE}}=3000$. 

\subsection{Effect on Stability}

\begin{figure}
\centering{}\includegraphics[width=0.4\paperwidth]{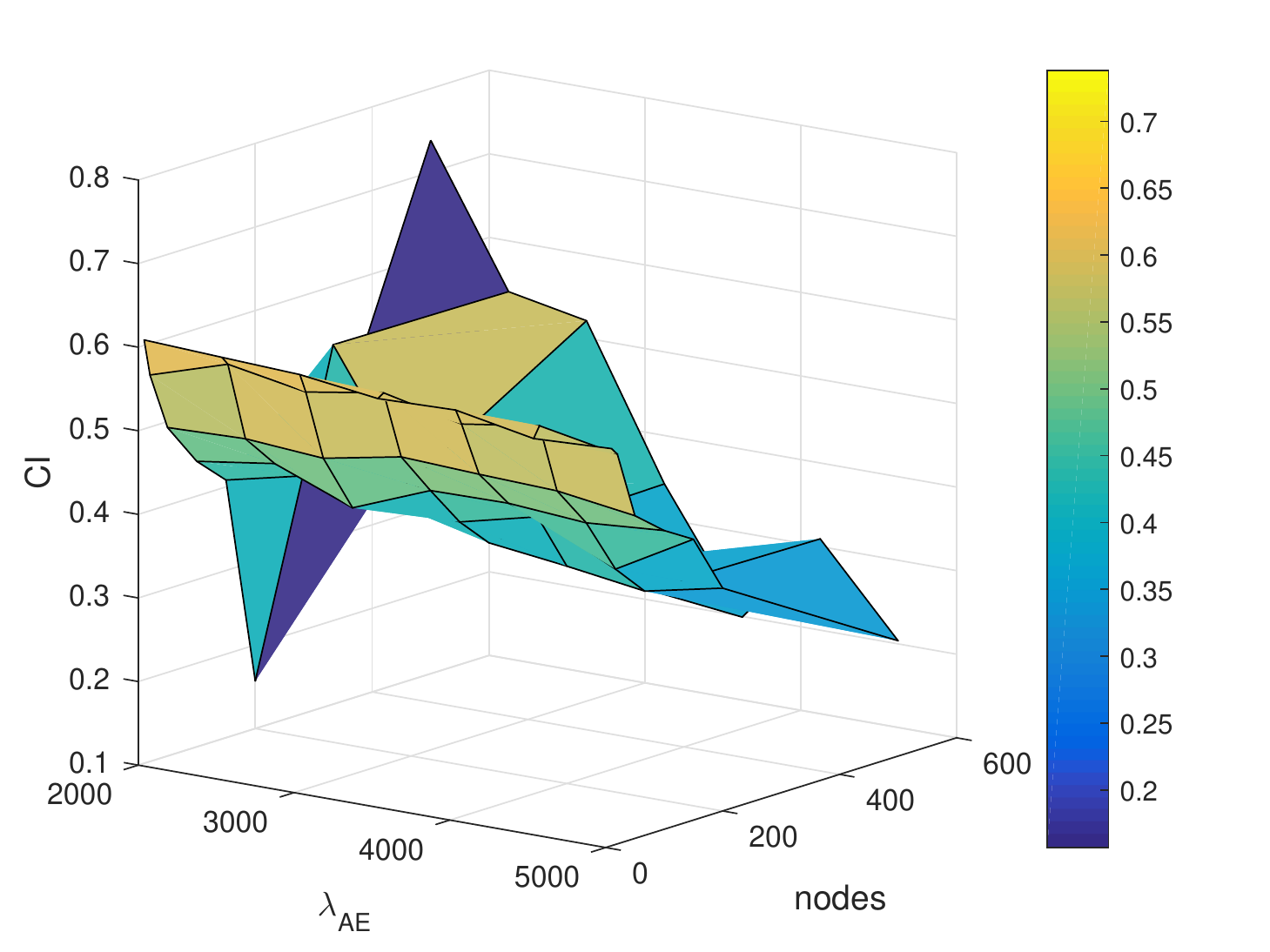}\caption{Effect of number of hidden units (nodes) and autoencoder penalty ($\lambda_{\text{AE}}$)feature
stability measured by consistency of top 100 features. Lasso parameter
fixed at $\alpha=.005$ \label{fig:hyperparams-stability}}
\end{figure}

\begin{figure}
\begin{centering}
\includegraphics[width=0.4\paperwidth]{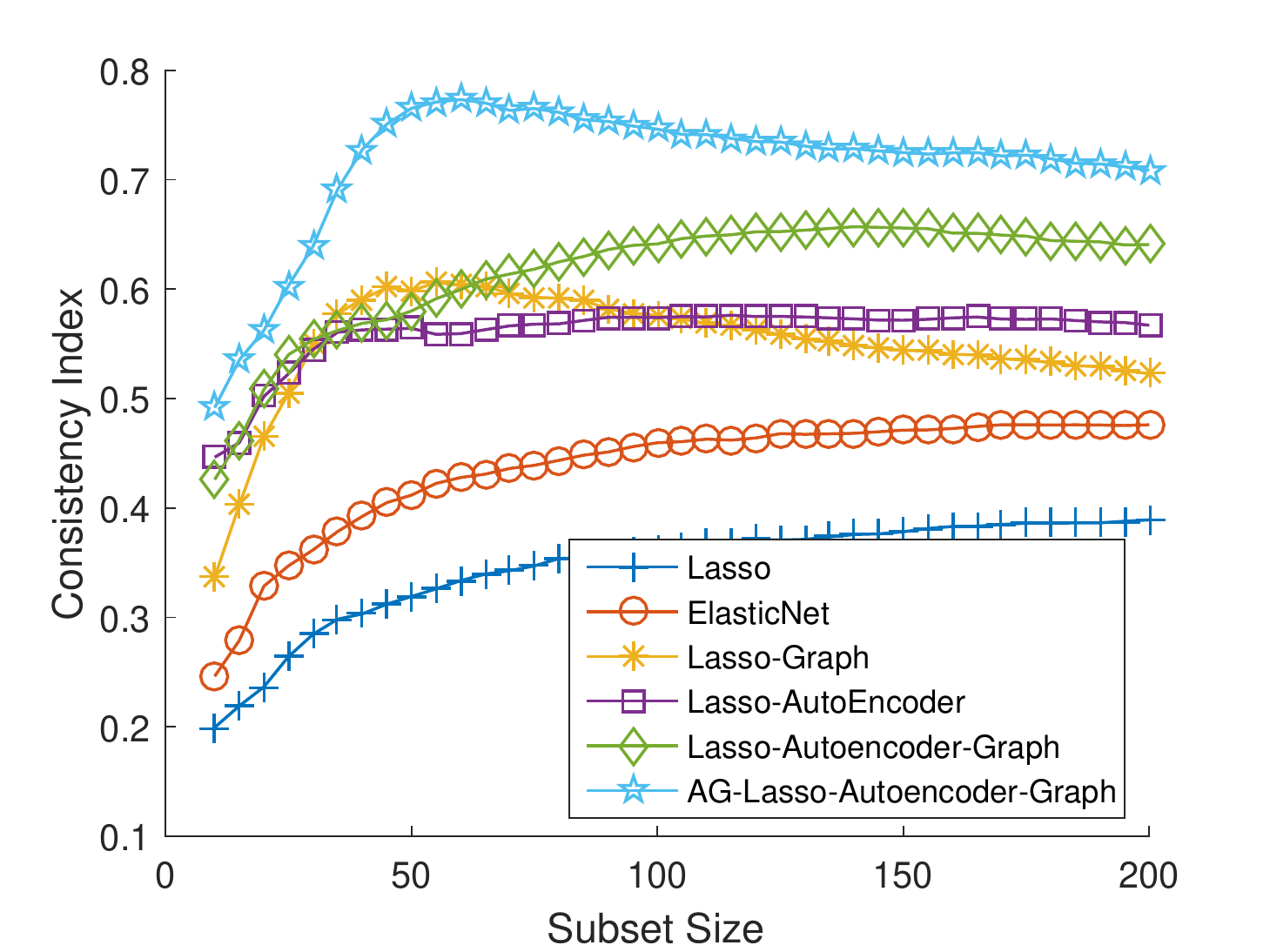}
\par\end{centering}
\caption{Feature stability as measured using Consistency Index. The plot compares
similarity in feature subsets generated by our proposed models and
baselines. Higher values indicate more stability.\label{fig:Feature-stability}}
\end{figure}

\begin{figure}
\centering{}\includegraphics[width=0.4\paperwidth]{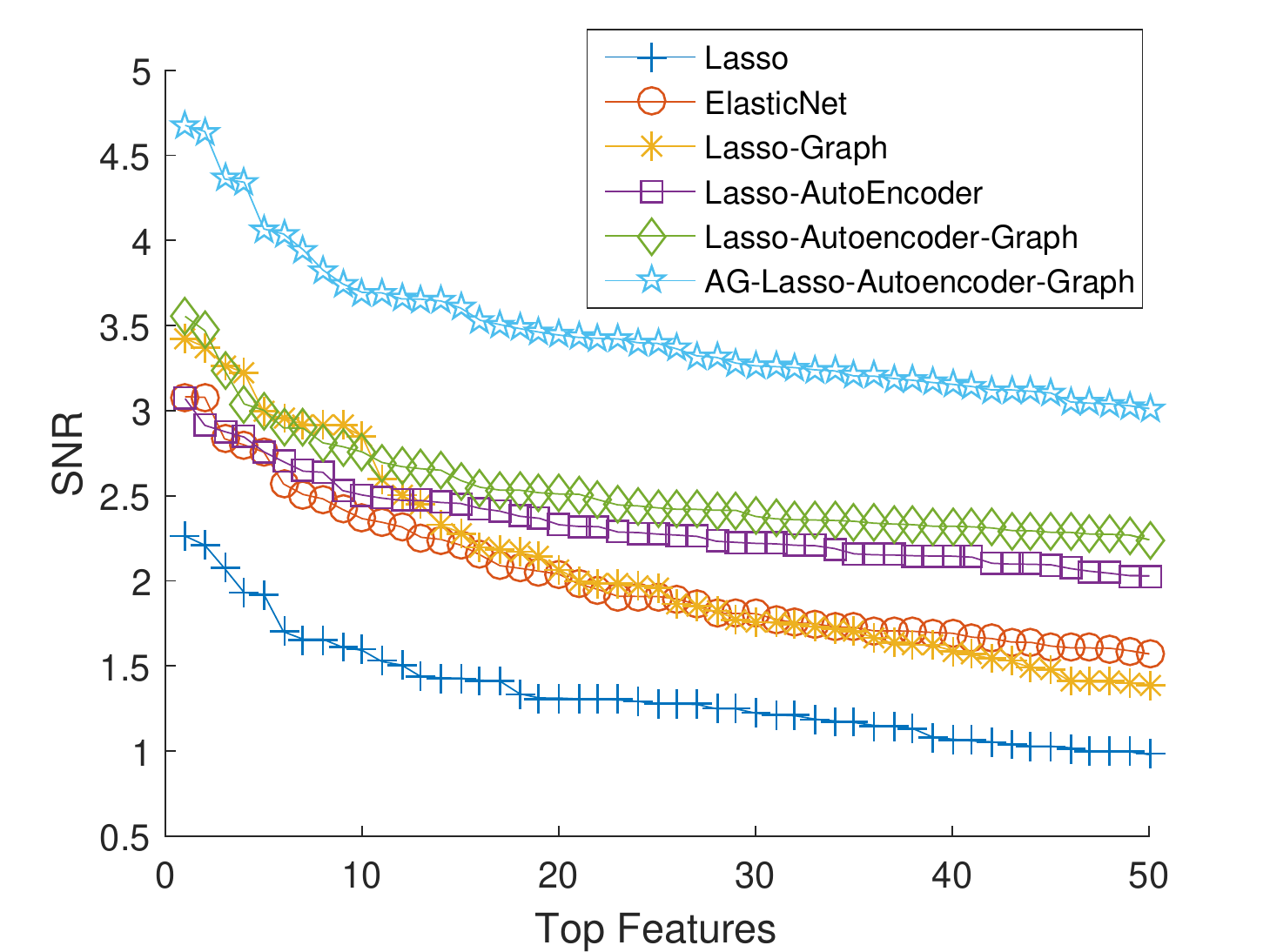}\caption{Model stability as measured using signal-to-noise ratio (SNR) of feature
weights. Higher values indicate more stability. \label{fig:Model-stability}}
\end{figure}

When compared to $\lambda_{\text{AE}}$, the choice of hidden units
had more influence on feature stability (see Fig.~\ref{fig:hyperparams-stability}).
 Consistency index measurements for feature selection stability is
reported in Fig.~\ref{fig:Feature-stability}.
 In general, capturing higher order correlations using autoencoder
improved feature stability when compared to baselines. Even though
pure autoencoder regularization proved to be more effective for larger
feature sets (\textgreater{}~120), the combination of autoencoder
and graph regularization consistently outperformed the baselines.
Further, augmenting external cohort to autoencoder learning resulted
in the most stable features. Similar observations were made when measuring
model estimation stability. Fig.~\ref{fig:Model-stability} reports
the signal-to-noise ratios of the top 50 individual features. 
At 95\% CI (approximately 1.96 std), lasso regularization identifies
3 features, elastic net identifies: 21, Graph regularization: 24,
while the autoencoder regularized models identify all the 50 features.
The variance in feature weight is greatly reduced by AG-Lasso-Autoencoder-Graph
regularization.

\section{Discussion and Conclusion}
Sparsity and stability are two important characteristics of interpretable
healthcare. Sparsity promotes interpretability and stability inspires
confidence in the prediction model. Through our experiments, we have
demonstrated that autoencoder regularization, when applied to high-dimensional
clinical prediction results in a sparse model that is stable in features
and estimation. Our proposed model is built from common clinical and
administrative data recorded in the hospital database, and hence can
be easily integrated into existing systems. 

The predictive performance of our proposed model (as measured by AUC)
is comparable with existing studies \cite{Betihavas2012}. Our stabilization
scheme did not improve classification performance. A similar observation
was made by Kalousis et al. in their study on high dimensional feature
selection stability \cite{Kalousis2007}. However, as their study
noted, stable models impart confidence on the features selected, and
in turn lends credence to corresponding classification performance. 

\subsection{Conclusion}

Traditionally, autoencoder variants are used to improve prediction/classification
accuracy. In this paper, we demonstrate a novel use of autoencoders
to stabilize high dimensional clinical prediction. Feature stable
models are reproducible between model updates. Stable models encourage
clinicians to further analyse the predictors in understanding the
prognosis, thereby paving the way for interpretable healthcare. We
have demonstrated that the encoding process of an autoencoder, though
intrinsically unstable, can be applied to regularize sparse linear
prediction resulting in more stable features.  The encoding weights
capture higher level correlations in EMR data. When collecting data
becomes expensive, augmenting another cohort during autoencoder training
resulted in a more robust estimation of encoding weights, translating
to better stability. This approach belongs to the emerging learning
paradigm of self-taught learning \cite{raina2007self}. We believe
this work presents interesting possibilities in the application of
deep nets for model stability.


\begin{thebibliography}{10}
\providecommand{\url}[1]{\texttt{#1}}
\providecommand{\urlprefix}{URL }

\bibitem{au2005attribute}
Au, W.H., Chan, K.C., Wong, A.K., Wang, Y.: Attribute clustering for grouping,
  selection, and classification of gene expression data. IEEE/ACM Transactions
  on Computational Biology and Bioinformatics (TCBB)  2(2),  83--101 (2005)

\bibitem{Austin2004a}
Austin, P.C., Tu, J.V.: Automated variable selection methods for logistic
  regression produced unstable models for predicting acute myocardial
  infarction mortality. Journal of clinical epidemiology  57(11),  1138--1146
  (2004)

\bibitem{bengio2009learning}
Bengio, Y.: Learning deep architectures for {AI}. Foundations and trends in
  Machine Learning  2(1),  1--127 (2009)

\bibitem{Betihavas2012}
Betihavas, V., Davidson, P.M., Newton, P.J., Frost, S.a., Macdonald, P.S.,
  Stewart, S.: What are the factors in risk prediction models for
  rehospitalisation for adults with chronic heart failure? Australian critical
  care : official journal of the Confederation of Australian Critical Care
  Nurses  25(1),  31--40 (Feb 2012),
  \url{http://www.ncbi.nlm.nih.gov/pubmed/21889893}

\bibitem{Cun2013}
Cun, Y., Fr{\"o}hlich, H.: Network and data integration for biomarker signature
  discovery via network smoothed t-statistics. PloS one  8(9),  e73074 (2013)

\bibitem{gopakumar_et_al_jbhi}
Gopakumar, S., Tran, T., Nguyen, T.D., Phung, D., Venkatesh, S.: Stabilizing
  highdimensional prediction models using feature graphs. IEEE Journal of
  Biomedical and Health Informatics  19(3),  1044--1052 (2015)

\bibitem{jacob2009group}
Jacob, L., Obozinski, G., Vert, J.P.: Group lasso with overlap and graph lasso.
  In: Proceedings of the 26th annual international conference on machine
  learning. pp. 433--440. ACM (2009)

\bibitem{Kalousis2007}
Kalousis, A., Prados, J., Hilario, M.: Stability of feature selection
  algorithms: a study on high-dimensional spaces. Knowledge and information
  systems  12(1),  95--116 (2007)

\bibitem{kamkar2015exploiting}
Kamkar, I., Gupta, S.K., Phung, D., Venkatesh, S.: Exploiting feature
  relationships towards stable feature selection. In: Data Science and Advanced
  Analytics (DSAA), 2015. 36678 2015. IEEE International Conference on. pp.
  1--10. IEEE (2015)

\bibitem{Kuncheva2007}
Kuncheva, L.I.: A stability index for feature selection. In: Artificial
  Intelligence and Applications. pp. 421--427 (2007)

\bibitem{Li2008}
Li, C., Li, H.: Network-constrained regularization and variable selection for
  analysis of genomic data. Bioinformatics  24(9),  1175--1182 (2008)

\bibitem{lin2013high}
Lin, W., Lv, J.: High-dimensional sparse additive hazards regression. Journal
  of the American Statistical Association  108(501),  247--264 (2013)

\bibitem{ma2007supervised}
Ma, S., Song, X., Huang, J.: Supervised group lasso with applications to
  microarray data analysis. BMC bioinformatics  8(1),  1--17 (2007)

\bibitem{Meinshausen2010}
Meinshausen, N., B{\"u}hlmann, P.: Stability selection. Journal of the Royal
  Statistical Society: Series B (Statistical Methodology)  72(4),  417--473
  (2010)

\bibitem{park2007averaged}
Park, M.Y., Hastie, T., Tibshirani, R.: Averaged gene expressions for
  regression. Biostatistics  8(2),  212--227 (2007)

\bibitem{raghupathi2014big}
Raghupathi, W., Raghupathi, V.: Big data analytics in healthcare: promise and
  potential. Health Information Science and Systems  2(1),  1--10 (2014)

\bibitem{raina2007self}
Raina, R., Battle, A., Lee, H., Packer, B., Ng, A.Y.: Self-taught learning:
  transfer learning from unlabeled data. In: Proceedings of the 24th
  international conference on Machine learning. pp. 759--766. ACM (2007)

\bibitem{Sandler2009}
Sandler, T., Blitzer, J., Talukdar, P.P., Ungar, L.H.: Regularized learning
  with networks of features. In: Advances in Neural Information Processing
  Systems 21, pp. 1401--1408. Curran Associates, Inc. (2009)

\bibitem{simon2011regularization}
Simon, N., Friedman, J., Hastie, T., Tibshirani, R., et~al.: Regularization
  paths for cox's proportional hazards model via coordinate descent. Journal of
  statistical software  39(5),  1--13 (2011)

\bibitem{Tibshirani1996}
Tibshirani, R.: Regression shrinkage and selection via the lasso. Journal of
  the Royal Statistical Society. Series B (Methodological) pp. 267--288 (1996)

\bibitem{Tibshirani2005}
Tibshirani, R., Saunders, M., Rosset, S., Zhu, J., Knight, K.: Sparsity and
  smoothness via the fused lasso. Journal of the Royal Statistical Society:
  Series B (Statistical Methodology)  67(1),  91--108 (2005)

\bibitem{tran2013integrated}
Tran, T., Phung, D., Luo, W., Harvey, R., Berk, M., Venkatesh, S.: An
  integrated framework for suicide risk prediction. In: 19th ACM SIGKDD
  international conference on Knowledge discovery and data mining. pp.
  1410--1418. ACM (2013)

\bibitem{Truyen2014ordinal_kais}
Tran, T., Phung, D., Luo, W., Venkatesh, S.: Stabilized sparse ordinal
  regression for medical risk stratification. Knowledge and Information Systems
  pp. 1--28 (2014)

\bibitem{Ye2012a}
Ye, J., Liu, J.: Sparse methods for biomedical data. ACM SIGKDD Explorations
  Newsletter  14(1),  4--15 (2012)

\bibitem{Yu2008}
Yu, L., Ding, C., Loscalzo, S.: Stable feature selection via dense feature
  groups. In: Proceedings of the 14th ACM SIGKDD international conference on
  Knowledge discovery and data mining. pp. 803--811. ACM (2008)

\bibitem{Yuan2006}
Yuan, M., Lin, Y.: Model selection and estimation in regression with grouped
  variables. Journal of the Royal Statistical Society: Series B (Statistical
  Methodology)  68(1),  49--67 (2006)

\bibitem{zhao2006model}
Zhao, P., Yu, B.: On model selection consistency of lasso. The Journal of
  Machine Learning Research  7,  2541--2563 (2006)

\bibitem{zhou2013patient}
Zhou, J., Sun, J., Liu, Y., Hu, J., Ye, J.: Patient risk prediction model via
  top-k stability selection. In: In Proceedings of the 13th SIAM International
  Conference on Data Mining. SIAM (2013)

\bibitem{zou2005regularization}
Zou, H., Hastie, T.: Regularization and variable selection via the elastic net.
  Journal of the Royal Statistical Society, Series B  67,  301--320 (2005)

\end{thebibliography}
\end{document}